\documentclass{article}

\usepackage[preprint,nonatbib]{neurips_2020}

\usepackage[utf8]{inputenc} % allow utf-8 input
\usepackage[T1]{fontenc}    % use 8-bit T1 fonts
\usepackage{hyperref}       % hyperlinks
\usepackage{url}            % simple URL typesetting
\usepackage{booktabs}       % professional-quality tables
\usepackage{amsfonts}       % blackboard math symbols
\usepackage{nicefrac}       % compact symbols for 1/2, etc.
\usepackage{microtype}      % microtypography
\usepackage{color}

\usepackage[title]{appendix}    % for appendix title

\usepackage{algorithm,algcompatible,amsmath}
\usepackage{algpseudocode}
\usepackage{multicol}
\usepackage{multirow}
\usepackage{graphicx}
\usepackage[caption=false,font=footnotesize]{subfig}
\newcommand{\deflen}[2]{%
    \expandafter\newlength\csname #1\endcsname
    \expandafter\setlength\csname #1\endcsname{#2}%
}

\title{Gradient Regularized Contrastive Learning for Continual Domain Adaptation}
\author{%
  Peng Su\footnotemark[1] \footnotemark[2],
    Shixiang Tang\footnotemark[2],
    Peng Gao\footnotemark[1],
    Di Qiu \footnotemark[1],
    Ni Zhao \footnotemark[1],
   Xiaogang Wang\footnotemark[1] \\
  The Chinese University of Hong Kong \footnotemark[1] \\
  \texttt{\{psu,xgwang\}@ee.cuhk.edu.hk} \\
  SenseTime Research \footnotemark[2] \\
}

\begin{document}

\maketitle

\begin{abstract}

Human beings can quickly adapt to environmental changes by leveraging learning experience.
However, the poor ability of adapting to dynamic environments remains a major challenge for AI models.
To better understand this issue, we study the problem of continual domain adaptation, where the model is presented with a labeled source domain and a sequence of unlabeled target domains.
There are two major obstacles in this problem: domain shifts and catastrophic forgetting.
In this work, we propose Gradient Regularized Contrastive Learning to solve the above obstacles.
At the core of our method, gradient regularization plays two key roles: (1) enforces the gradient of contrastive loss not to increase the supervised training loss on the source domain, which maintains the discriminative power of learned features; (2) regularizes the gradient update on the new domain not to increase the classification loss on the old target domains, which enables the model to adapt to an in-coming target domain while preserving the performance of previously observed domains.
Hence our method can jointly learn both semantically discriminative and domain-invariant features with labeled source domain and unlabeled target domains.
The experiments on Digits, DomainNet and Office-Caltech benchmarks demonstrate the strong performance of our approach when compared to the state-of-the-art.

\end{abstract}

\section{Introduction}
Deep learning has shown great generalization power on various benchmarks when the training and testing data are drawn from the same distribution.
However, the model trained on an existing benchmark cannot generalize effectively to a new scenario due to the well-known domain shift problem. The existence of domain shift hinders the deployment of deep learning models in the open environment of the real-world. Extensive Domain Adaptation (DA) methods~\cite{ganin2016domain,long2017deep,tzeng2017adversarial,saito2018maximum,su2020adapting} have been proposed to enable models to generalize from a label-rich source domain to an unlabeled target domain.
Most of the literature of domain adaptation focus on adapting from source domain to one target domain~\cite{ganin2016domain,long2017deep,tzeng2017adversarial,saito2018maximum,su2020adapting,peng2019domain} or multiple target domains~\cite{gong2013reshaping,gholami2020unsupervised,liu2019compound}.
However, many machine learning models deployed in the real-world are exposed to non-stationary situations where different domains are acquired sequentially and their distribution varies over time, such as deep learning models in autonomous vehicles.
When meeting the scenario where target domains are multiple and are coming sequentially, existing DA methods collapse because the models may quickly adapt to new domains and easily forget knowledge on old target domains. The killer defect, which is known as \emph{catastrophic forgetting}, prevents most DA algorithms being put into practice.

This work studies the problem of continual domain adaptation that the models are required to continually adapt to new target domains without harming performances on the previously observed domains.
Previous continual DA methods  ~\cite{bobu2018adapting,mancini2019adagraph} require priors about domain labels to conduct domain adversarial training or build a graph of inter-domain relationships. However these priors may not be accessible in practice, i.e. the target domain is compounded with multiple subdomains.
Besides, the existing method~\cite{bobu2018adapting} applies naive sample replay to avoid catastrophic forgetting, which still suffers from the large negative backward transfer.

In this work, we propose gradient regularized contrastive learning (GRCL) to tackle ``domain shifts'' and ``catastrophic forgetting'' in a unified framework.
At the core of GRCL, gradient regularization plays two key roles: (1) enforces the gradient of contrastive loss not to increase the supervised training loss on the source domain, which maintains the discriminative power of source features, and in turn improves the target features learned by contrastive loss; (2) regularizes the gradient update on the new domain not to increase the classification loss on the domain memory, which enables the model to adapt to an in-coming target domain while preserving the performance of previously observed domains.
Hence GRCL can jointly learn semantically discriminative and domain-invariant features, without reliance on priors of domain-related information.

\begin{figure}[!t]
    \centering
    \includegraphics[width=1\linewidth]{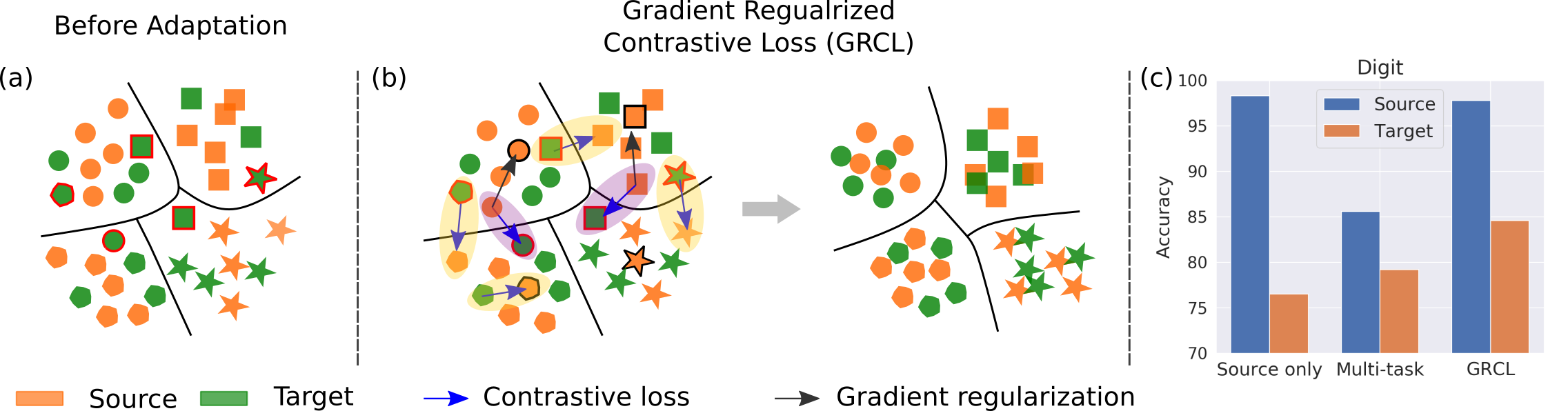}
    \caption{Illustration of GRCL. In Fig.(a)(b) the shape denotes class and color denotes domain.
    The contrastive loss pushes target samples towards its similar ones in the source domain.
    However, it inevitably pushes some discriminative features towards the instance features of lower discriminative power (purple area), as the contrastive loss only captures the visual similarity while ignoring the semantics.
    Such a problem is verified by the severe performance degradation on the source domain (Fig.(c)).
    GRCL regularizes the model update not to increase the loss on source domain (black arrow), that maintains the discriminative power of learned features.
    }
    \label{fig:demo-grcl}
\end{figure}

Specifically,  we construct a  domain memory to store a subset of image samples from each domain.
GRCL leverages the contrastive loss to jointly learn image representation with the domain memory and the incoming target domain.
Because the instances belong to the same category exhibit similar appearances, the contrast loss encourages the model to push the target samples towards the samples in the domain memory that belong to the same class, in feature space.
Thus different domain inputs belong to the same category will be aligned in the feature space, resulting in domain-invariant image representation.
However, simply combining the contrastive learning and supervised learning (with source domain data) as a multi-task learning objective could hurt the discriminative power of learned features, due to the conflict between contrastive loss and cross-entropy loss.
As shown in the purple area of Fig.\ref{fig:demo-grcl}(b), the contrastive loss inevitably pushes some discriminative features towards the instance features of less discriminative power, as the contrastive loss only captures the visual similarity while ignoring the semantics.
Such a problem is empirically verified by the experimental results in section~\ref{sec:grcl-exp}.
To solve this conflict, we enforce the gradient of contrastive loss not to increase the cross-entropy loss on the source domain, which maintains the discriminative power of source features. Since the source instance features act as the anchors for the target instances in contrastive learning, it also improves the quality of learned target features.
To overcome ``catastrophic forgetting'', we construct an additional set of constraints in which each constraint is imposed to enforce the classification loss of each domain-specific memory never increasing.

To summarize, our contributions are as follows:
(1) We propose gradient regularized contrastive learning to jointly learn both discriminative and domain-invariant representation without reliance on priors of domain labels. At the core of our method, gradient regularization performs two key functions: maintaining the semantically discriminative power of learned features and overcoming catastrophic forgetting.
(2) The experiments on multiple continue domain adaptation benchmarks demonstrate the strong performance of our method when compared to the state-of-the-art.

\section{Problem Formulation and Evaluation Metric}

Let $\mathcal{D}_s = \{ (x_i^s, y_i^s) \}_{i=1}^{n_s}$ be the labeled dataset of source domain, where each example $(x_i^s, y_i^s)$ is composed of an image $x_i^s \in \mathcal{X}^s$ and its label $y_i^s \in \mathcal{Y}$.
Continue domain adaptation defines a sequence of adaptation tasks $\mathcal{T} = \{ \mathcal{T}_1, \mathcal{T}_2, \dots, \mathcal{T}_N \}$.
Different domains have a common label space $\mathcal{Y}$ but distinct data distributions.
On $t$-th task $\mathcal{T}_t$, there is an {\it unlabeled} target domain dataset $\mathcal{D}_t =\{ x_i^t\}_{i=1}^{n_t}$.
The goal is to learn a label prediction model $f$ that can generalize well on multiple target domains $\{\mathcal{D}_1, \dots, \mathcal{D}_N \}$. Note that data from different domains in general follow different distributions.

We propose two metrics to evaluate the model adapting over a stream of target domains, namely average accuracy (ACC) and average backward transfer (BWT).
After the model finishes the training of adaptation task $\mathcal{T}_i$, we evaluate its individual performance on the testing set of the current and all the previously observed domains $\mathcal{D}_k^{test} (\forall k \leq i)$.
Let $R_{i, j}$ denote the test accuracy of the model on the domain $\mathcal{D}_j$ after finishing adapting to domain $\mathcal{D}_i$. We use $\mathcal{D}_0$  to denote the source domain.
ACC and BCT can be calculated as
\begin{equation*}
     ACC  = \frac{1}{N} \sum_{i=0}^{N} R_{N, i}   \qquad
     BWT  = \frac{1}{N-1} \sum_{i=1}^{N-1} R_{N, i} - R_{i, i}.
\end{equation*}
The ACC represents the average performance over all domains when the model has finished the last adaptation task.
And BCT indicates the influence that adapting to domain $\mathcal{D}_t$ has on the performance on a previously observed domain $\mathcal{D}_{k < t}$.
The negative BWT indicates that adapting to a new domain decreases the performance on previous domains.
The larger these two metrics, the better the model.

\section{Method}

\subsection{Gradient Regularized Contrastive Learning} \label{sec:3-1}

Contrastive learning~\cite{wu2018unsupervised,he2019momentum,chen2020simple} has recently shown the great capability of mapping images to an embedding space, where similar images are close together and dissimilar images are far apart.
Inspired by this, we utilize the contrastive loss to push the target instance towards the source instances that own similar appearances with target input.
Thus the same category instances that are apparently similar but from different domains are aligned in the feature space, resulting in domain-invariant features.
Specifically, we first define an episodic memory $\mathcal{M}_t $, which stores a subset of observed images from target domain $\mathcal{D}_t $.
When the model is adapting to $t$-th domain, we have a domain memory $\mathcal{M} = \cup_{i=1}^{t-1} \mathcal{M}_i $, that is a union of all past episodic memories.

\begin{figure}[]
    \centering
    \includegraphics[width=0.9\linewidth]{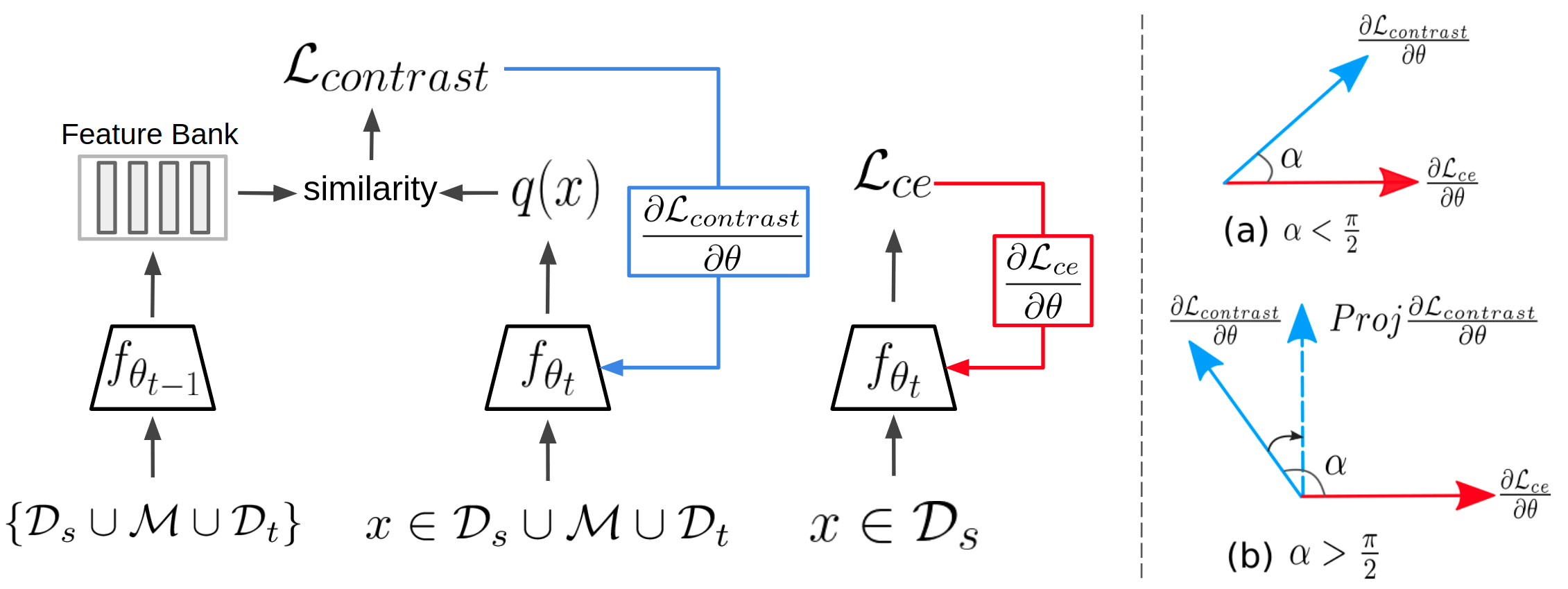}
    \caption{Gradient Regularized Contrastive Learning.
    The gradient regularization (right) is utilized to enforce the  gradient of contrastive loss not to increase the cross-entropy loss, which maintains the discriminative power of feature learned by contrastive learning.
    }
    \label{fig:fig2}
\end{figure}

Let $f_{\theta_{t-1}}$ be the model finishing the adaptation training on domain $\mathcal{D}_{t-1}$.
We construct a domain-affiliate feature bank $\mathcal{B}_t=\{ k(x), \forall x \in  \mathcal{D}_s \cup \mathcal{M} \cup \mathcal{D}_t \}$ to store the instance features from both source and various target domains (see Fig.~\ref{fig:fig2}). Here, $k(x)$ is a compact representation of the input $x$, which is initialized by $k(x) = g(f(x, \theta_{t-1}))$ and normalized so that $\| k(x) \|_2^2 =1$. We choose
$g(\cdot)$ to be a 2-layer MLP $g(z) = W_2 \sigma (W_1 z)$ that maps the semantic features to a low-dimensional vector of $dim = 128$.
Given an query input $q = k(x)$, we assume there is a single positive key $k_+$ that $q$ matches.
$k_+$ is designed as $ g(f(T(x), \theta))$, where $T(\cdot)$ denotes the data augmentation.
With similarity measured by dot product, we consider the contrastive loss function of InfoNCE \cite{oord2018representation} :
\begin{equation} \label{eq:2}
\mathcal{L}_{q(x), k^+, \{k^{-}\}} = -\log \frac{\exp (q \cdot k^+ / \tau)}{\exp (q \cdot k^+ / \tau) + \sum\limits_{k^{-}\in   \mathcal{B}_t} \exp(q \cdot k^{-} / \tau)},
\end{equation}
where  $k^{-}$ represents the negative key for $q$, and $\tau$ is the temperature.
During the training, we update the sampled keys with the up-to-date model $f_{\theta_t}$ by $k \xleftarrow{} m \cdot k^{i-1} + (1-m) k^i$, which $i$ indicates the training iteration and $m$ is the momentum.
Then the adaptation objective becomes a multi-task of supervised training loss on $\mathcal{D}_s$ and contrastive learning loss on the feature bank $\mathcal{B}_t$:

\begin{equation} \label{eq:contrast-da-multi-task}
    \begin{aligned}
    \min_{\theta_t} \quad \mathcal{L} & = \mathcal{L}_{ce}(\theta_t, \mathcal{D}_s) + \lambda \mathcal{L}_{contrast}(\theta_t, \mathcal{B}_t) \\
     & =  \frac{1}{| b_s |} \sum_{(x,y) \in \mathcal{D}_s }CE(f(x; \theta_t), y) +
    \lambda \frac{1}{|b_c |} \sum_{x \in \mathcal{D}_s \cup \mathcal{M} \cup \mathcal{D}_t  }^{} \mathcal{L}_{q(x), k^+, \{k^{-}\}},
    \end{aligned}
\end{equation}
where $\mathcal{L}_{ce}$ is the cross entropy loss on source domain $\mathcal{D}_s $.
$b_s$, $b_t$ denote the mini-batch of samples from  $\mathcal{D}_s $ and union of $\mathcal{D}_s \cup \mathcal{M} \cup \mathcal{D}_t$ respectively.
$\lambda$ is the hyper-parameter to trade off the two losses.

However, our experiments show that the multi-task objective of Eq.\ref{eq:contrast-da-multi-task} only brings marginal improvements on the target domain, as shown in Fig.\ref{fig:demo-grcl} (c).
We hypothesize the problem raised from the conflict of two objectives, namely cross-entropy loss  and contrastive loss.
As illustrated in the purple area in Fig.\ref{fig:demo-grcl}(b), the discriminative instance features  could be pushed towards less-discriminative instance features, because of the apparent similarity.
This conflict is verified by the experiments that contrast loss degrades the model performance on the source domain (see Section~\ref{sec:grcl-exp}), suggesting hurting the discriminative power of image features from the source domain.
Since the source instance features act as the anchors for the target instances in contrastive learning, the conflict may impose an undesirable influence on the discriminative power of image features from the target domain.

To solve the conflict between $\mathcal{L}_{contrast}$ and $\mathcal{L}_{ce}$, we enforce the regularization that the $\mathcal{L}_{ce}$ on the source domain should not increase when minimizing the $\mathcal{L}_{contrast}$.
Then the final domain adaptation objective with constraints can be rephrased as:

\begin{equation} \label{eq:grcl}
           \min_{ \theta_t} \quad  \mathcal{L}_{contrast}(\theta_t, \mathcal{B}_t) \qquad
    \text{subject to} \quad
     \langle g_s,  g_t \rangle \geq 0 ,
\end{equation}
where $\langle g_t,  g_s \rangle  $ is the inner product of gradient of loss $\mathcal{L}_{contrast}$ and $\mathcal{L}_{ce}$ w.r.t. model parameters:
\begin{equation} \label{eq:4}
 \langle g_s, g_{t} \rangle := \langle \frac{\partial \mathcal{L}_{ce}(\theta_t, \mathcal{D}_s)}{ \partial \theta_t},   \frac{\partial \mathcal{L}_{contrast}(\theta_t, \mathcal{B}_t) }{ \partial \theta_t} \rangle \geq 0.
\end{equation}
And $\theta_t$ is initialized with the model trained on the source domain with labels.
As illustrated in Fig.\ref{fig:fig2} (right),  if the constraint of Eq.\ref{eq:4} is satisfied, then the parameter update $g_t$ is unlikely to hurt the discriminative power of image features from the source domain, as it does not increase the cross-entropy loss on the source domain.
If the violations occur, we propose to project the $g_t$ to the closest gradient $ \hat g_t$ satisfying the constraint of Eq.\ref{eq:4}.
Thus GRCL enjoys the benefits of semantically discriminative features offered by the gradient regularization and domain-invariant features learned by contrastive loss.

\subsection{Overcoming Catastrophic Forgetting} \label{sec:3-2}
In this section, we extend the GRCL with additional constraints to overcome "catastrophic forgetting", in which each constraint is imposed to enforce the classification loss of each domain-specific memory never increasing.
Mathematically, the constraints can be formulated as:
\begin{equation} \label{eq:6-0}
 \mathcal{L}_{ce}(\theta_t, \mathcal{M}_k) \leq \mathcal{L}_{ce}(\theta_{t-1}, \mathcal{M}_k )\quad \text{for all} \quad k < t.
\end{equation}
where $\theta_{t-1}$ is the model parameters at the end of adapting task on $\mathcal{D}_{t-1}$.
While Eq.\ref{eq:6-0} effectively permits positive backward transfer of GRCL, it comes at a huge computation burden at training time.
At each training step, we need to solve the $t-1$ inequality constraints of all episodic memories.
It will become prohibitive when the size of $\mathcal{M}_k$ and number of adaptation tasks are large.
Alternatively, we propose a much efficient way to approximate the Eq.\ref{eq:6-0} by

\begin{equation} \label{eq:6}
 \mathcal{L}_{ce}(\theta_t, \mathcal{M}) \leq \mathcal{L}_{ce}(\theta_{t-1}, \mathcal{M} ),
\end{equation}
where  $\mathcal{M} = \cup_{i=1}^{t-1} \mathcal{M}_i $ is the domain memory.
Instead of computing the loss on each individual old domain, Eq.\ref{eq:6} only computes the loss with the sampled batch of images from domain memory $\mathcal{M}$.
% how to get pseudo labels.
For computing $\mathcal{L}_{ce}(\theta_t, \mathcal{M})$, we adopt standard $k$-means clustering algorithm~\cite{caron2018deep} to generate pseudo labels with a pre-trained model obtained from previous domain adaptation task.
Combining the adaptation objective of Eq.\ref{eq:grcl} and against-forgetting objective of Eq.\ref{eq:6}, we have the final objective for continue domain adaptation:
\begin{equation}
\begin{aligned} \label{eq:7}
    \min_{ \theta_t} \quad & \mathcal{L}_{contrast}(\theta_t, \mathcal{B}_t) \\
    \text{subject to} \quad
    & \langle g_t,  g_s \rangle \geq 0 \\
    & \mathcal{L}_{ce}(\theta_t, \mathcal{M}) \leq \mathcal{L}_{ce}(\theta_{t-1}, \mathcal{M} ).
\end{aligned}
\end{equation}
The first constraint is to facilitate the contrastive learning to learn discriminative features.
The second constraint ensures the average loss on previously observed domains does not increase, which enforces the model not to forget acquired knowledge on preceding domains.
Mathematically, we want to find the gradient update $\hat g_t$ satisfying:
\begin{equation}   \label{eq:8}
    \begin{aligned}
    \min_{\hat g_t} \quad & \frac{1}{2} \| g_t -  \hat g_t \|_2^2 \\
    \text{subject to} \quad  & \langle \hat g_t,  g_s \rangle \geq 0 \\
     & \langle \hat g_t, g_{dm} \rangle \geq 0,
    \end{aligned}
\end{equation}
where $g_{dm}$ is the gradient computed using a batch of random samples from the domain memory $\mathcal{M}$.
Eq.\ref{eq:8} is a quadratic program (QP) on $P$ variables (the number of parameters in the neural network), which could be measured in millions for deep learning models.
To solve Eq.\ref{eq:8} efficiently, we work in the dual space of Eq.\ref{eq:8} which results in much smaller QP with only $2$ variables:
\begin{equation} \label{eq:10}
    \min_{u} \quad   \frac{1}{2} u^{T}GG^{T}u + g_t^{T}G^{T}u  \quad
    \text{subject to} \quad   u \geq 0,
\end{equation}
where $G = -(g_s, g_{dm}) \in \mathbf{R}^{ 2 \times P}$ and we discard the constant term of $g_t^{T}g_t $.
The formal proof of Eq.\ref{eq:10} is provided in Appendix~\ref{a:proof}.
Once the solution $u^*$ to Eq.\ref{eq:10} is found, we can solve the  Eq.\ref{eq:8} with the gradient update of $\hat g_t = G^{T}u^* + g_t $.
Appendix~\ref{a:algo-grcl} summarizes the training protocol of GRCL.

\section{Experiments}

\subsection{Dataset and Methods}

\paragraph{Digits} includes five digits datasets (MNIST~\cite{lecun1998gradient}, MNIST-M~\cite{ganin2015unsupervised}, USPS~\cite{hull1994database}, SynNum~\cite{ganin2015unsupervised} and SVHN~\cite{netzer2011reading}).
Each domain has $7,500$ images for training and $1,500$ images for testing.
We consider a continual domain adaptation problem of SynNum $\xrightarrow{}$ MNIST $\xrightarrow{}$ MNIST-M   $\xrightarrow{}$  USPS $\xrightarrow{}$ SVHN.
\paragraph{DomainNet}~\cite{Peng_2019_ICCV} is one of  the largest domain adaptation datasets with approximately $0.6$ million images distributed among $345$ categories.
Each domain randomly selects $40,000$ images for training and $8,000$ images for testing.
Five different domains from DomainNet are used to build a continual domain adaptation task as Clipart $\xrightarrow{}$ Real $\xrightarrow{}$ Infograph $\xrightarrow{}$  Sketch  $\xrightarrow{}$ Painting.

\paragraph{Office-Caltech}~\cite{gong2012geodesic}
includes 10 common categories shared by Office-31~\cite{saenko2010adapting} and Caltech-256~\cite{griffin2007caltech} datasets.
Office-31 dataset contains three domains: DSLR, Amazon and WebCam, which represent the images that are collected in different environments respectively.
We consider a continual domain adaptation tasks of DSLR $\xrightarrow{}$  Amazon $\xrightarrow{}$  WebCam $\xrightarrow{}$  Caltech.

% \subsection{Baseline Methods}
We compare GRCL  with five alternatives, including (1) DANN~\cite{ganin2016domain}, a classic domain adversarial training based method; (2) MCD~\cite{saito2018maximum}, maximizing the classifier discrepancy to reduce domain gap; (3) DADA~\cite{peng2019domain}, disentangling the domain-specific features from category identity; (4) CUA~\cite{bobu2018adapting}, adopting an adversarial training based method ADDA~\cite{tzeng2017adversarial} to reduce the domain shift and a sample replay loss to avoid forgetting; (5) GRA, replacing the contrastive learning in GRCL with ADDA~\cite{tzeng2017adversarial}.
% For adaptation method using memory, CUA, GRA and GRCL use exactly the same size of episodic memory for each domain and same  $k$-means algorithm to generate pseudo labels.
% Implementation details are provided in Appendix~\ref{a:implementation}.

% We compare GRCL  with five alternatives:
% \begin{enumerate}
%     % \item Source represents the model supervised trained with source data only.

%     \item DANN~\cite{ganin2016domain} is a classic domain adversarial training based method.

%     \item MCD~\cite{saito2018maximum} maximizes the classifier discrepancy to align target domain distribution.

%     \item DADA~\cite{peng2019domain} disentangles the domain-specific features from category identity.

%     \item CUA~\cite{bobu2018adapting} adopts an adversarial training based method ADDA~\cite{tzeng2017adversarial} to reduce the domain gaps and a sample replay loss to avoid forgetting.

%     \item GRA, replacing the contrastive learning in GRCL with ADDA~\cite{tzeng2017adversarial}.

% \end{enumerate}

For fair comparison with previous state-of-the-art, we adopt LeNet-5~\cite{lecun1998gradient} on Digits, ResNet-50~\cite{he2016deep} on DomainNet, and ResNet-18~\cite{he2016deep} on Office-Caltech benchmarks.
Each domain has the same number of training and testing images.
For contrastive learning, we use a batch size of 256, feature update momentum of $0.5$, number of negatives as 1024, training schedule of 240 epochs.
The MLP head uses a hidden dimension of $2048$.
Following~\cite{wu2018unsupervised,he2019momentum}, the temperature $\tau$ in Eq.\ref{eq:2} is set as $0.07$.
For data augmentation, we use random color jittering, Gaussian blur and random horizontal flip.
And the image samples in the episodic memory are selected by the model predictions with top-1024 confidence.
For methods using memory, CUA, GRA and GRCL use exactly the same size of episodic memory for each domain and same  $k$-means algorithm to generate pseudo labels.

\subsection{Experimental Results}

\begin{figure}[!t]
    \centering
    \includegraphics[width=1\linewidth]{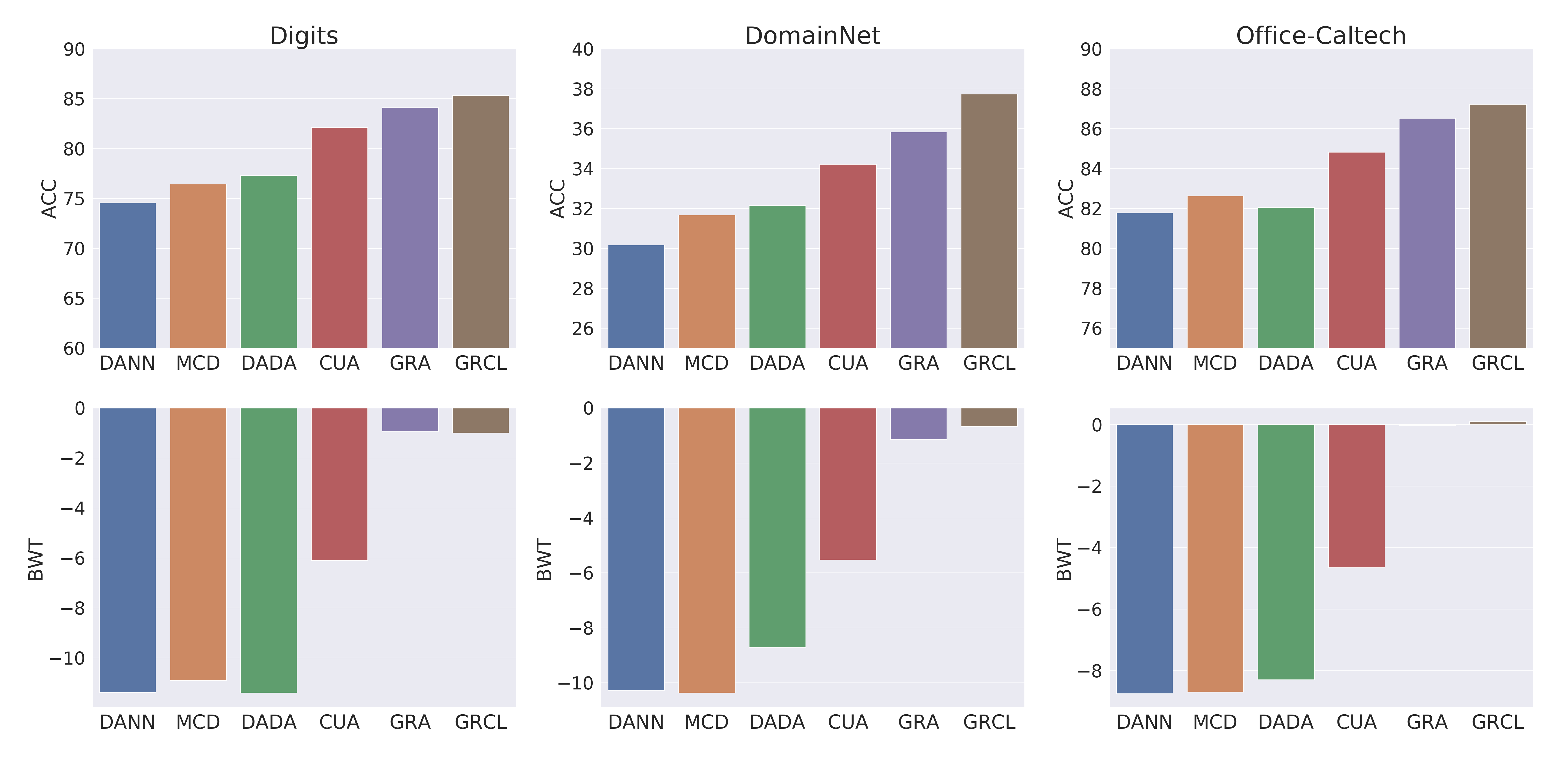}
    \caption{ACC and BWT on three continual domain adaptation benchmarks.}
    \label{fig:fig3-0}
\end{figure}

\begin{figure}[!h]
    \centering
    \includegraphics[width=1\linewidth]{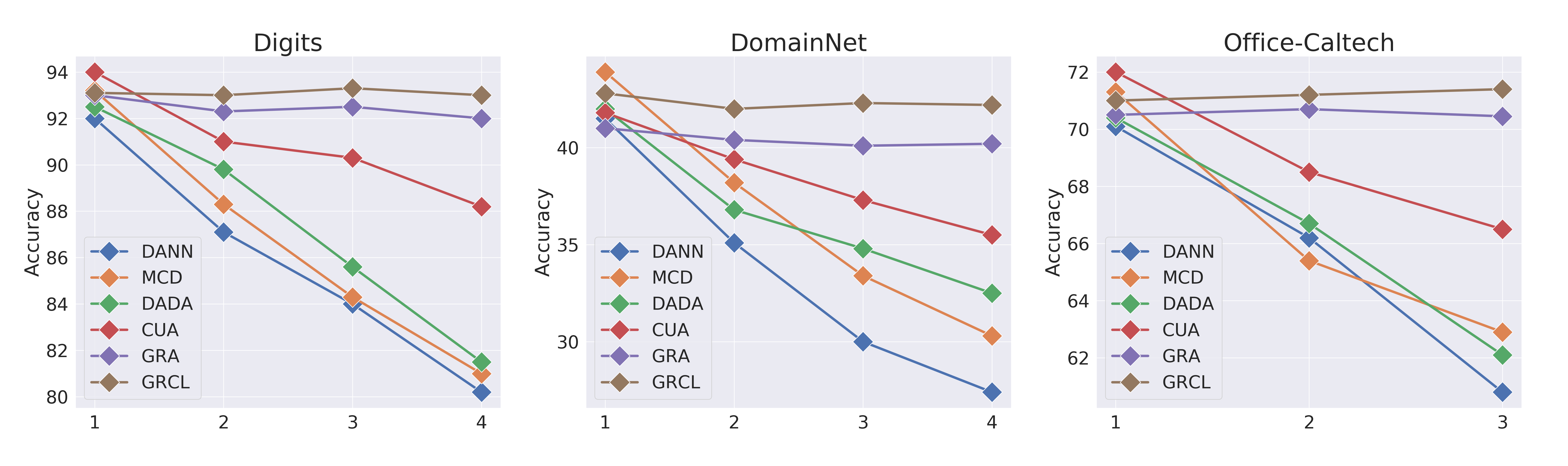}
    \caption{Evolution of classification accuracy on the first target domain as more domains are observed. Existing methods exhibit significant performance degradation due to catastrophic forgetting.}
    \label{fig:fig3}
\end{figure}

Fig.\ref{fig:fig3-0} shows the results on three benchmarks. The larger the average accuracy (ACC) and backward transfer (BWT) the better the model.
When the model has finished the training on the last target domain, we report the ACC over all observed domains.
As shown in Fig.\ref{fig:fig3-0}, GRCL consistently achieves better  ACC across three benchmarks, suggesting that the model trained with GRCL owns the best generalization capability across domains.
Unsurprisingly, most methods exhibit lower negative BWT, as catastrophic forgetting exists.
The methods using memory (CUA, GRA, GRCL) performs better than the other methods without memory (DANN, MCD, DADA), especially on the BWT metric.
These results highlight the importance of memory in the continual DA problem.

Among the memory-based methods, GRA and GRCL achieve significantly better BWT  on three benchmarks, suggesting the effectiveness of gradient constraints for combating catastrophic forgetting.
GRCL consistently achieves better ACC than GRA across all benchmarks.
It partially because that GRCL utilizes all the samples from domain memory (cached the samples from all previously observed domains) to jointly learn representation, while GRA only uses source domain and current target domain to learn features.
Fig.\ref{fig:fig3} depicts the evolution of classification accuracy on the first target domain as more domains are observed.
GRCL consistently exhibits minimal forgetting and even positive backward transfer on \textit{Office-Caltech} benchmark.
Table~\ref{tab:t1} summarizes the detailed results for all methods on three continual DA benchmarks.
Each entry in the Table~\ref{tab:t1} represents the mean and standard deviation of classification accuracy of five runs in corresponding experiments.

\begin{table}[]
\centering
\caption{ACC and BWT on three continue domain adaptation benchmarks.}
\label{tab:t1}
\begin{tabular}{@{}lcc|cc|cc@{}}
\toprule
Methods & \multicolumn{2}{c}{\textit{Digits}}                                                                       & \multicolumn{2}{c}{\textit{DomainNet}}                                     & \multicolumn{2}{c}{\textit{Office-Caltech}} \\ \midrule
                                    & ACC                               & BWT                                    & ACC                                  & BWT                                    & ACC                                   & BWT          \\ \cmidrule(l){2-3} \cmidrule(l){4-5} \cmidrule(l){6-7}
% Source                              & 72.76 \scriptsize $\pm$ 0.03       & -                          & 29.18 \scriptsize $\pm$ 0.05        & -                      & 80.03 \scriptsize $\pm$ 0.05        & -          \\

DANN~\cite{ganin2016domain}         & 74.56 \scriptsize $\pm$ 0.14      & -11.37 \scriptsize $\pm$ 0.09          & 30.18  \scriptsize $\pm$ 0.13        & -10.27  \scriptsize $\pm$ 0.07          & 81.78 \scriptsize $\pm$ 0.05        & -8.75    \scriptsize $\pm$ 0.07         \\
MCD~\cite{saito2018maximum}         & 76.46 \scriptsize $\pm$ 0.24       & -10.90 \scriptsize $\pm$ 0.11          & 31.68 \scriptsize $\pm$ 0.20        & -10.36  \scriptsize $\pm$ 0.15           & 82.63 \scriptsize $\pm$ 0.13        & -8.70   \scriptsize $\pm$ 0.12           \\
DADA~\cite{peng2019domain}          & 77.30 \scriptsize $\pm$ 0.19       & -11.40 \scriptsize $\pm$ 0.04          & 32.14  \scriptsize $\pm$ 0.14       & -8.67  \scriptsize $\pm$ 0.09           & 82.05 \scriptsize $\pm$ 0.03        & -8.30   \scriptsize $\pm$ 0.05           \\
CUA~\cite{bobu2018adapting}         & 82.12 \scriptsize $\pm$ 0.18        & -6.10 \scriptsize $\pm$ 0.12          & 34.22  \scriptsize $\pm$ 0.16       & -5.53  \scriptsize $\pm$ 0.14          & 84.83  \scriptsize $\pm$ 0.10       & -4.65   \scriptsize $\pm$ 0.08           \\ \hline
GRA                               & 84.10 \scriptsize $\pm$ 0.15        & -0.93 \scriptsize $\pm$ 0.10          & 35.84  \scriptsize $\pm$ 0.19       & -1.15  \scriptsize $\pm$ 0.16           & 86.53  \scriptsize $\pm$ 0.11       & -0.03   \scriptsize $\pm$ 0.03           \\
GRCL                                 & 85.34 \scriptsize $\pm$ 0.10        & -1.0 \scriptsize $\pm$ 0.03          & 37.74 \scriptsize $\pm$ 0.13        & -0.67  \scriptsize $\pm$ 0.12           & 87.23  \scriptsize $\pm$ 0.06       & 0.05      \scriptsize $\pm$ 0.02            \\ \bottomrule
\end{tabular}
\end{table}

\subsubsection{Importance of Memory Size and Training Schedule} \label{a:ablation-memory-size}

\begin{table}[!h]
  \resizebox{13.5cm}{!}{
    \begin{minipage}{.5\textwidth}
    % \fontsmall
        \caption{ACC as a function of memory size.}
        \label{tab:t2-1}
        \centering
            \begin{tabular}{ccccc}
            \toprule
            memory size             & 256        & 512       & 1024      & 2048 \\ \midrule
            \textit{Digits}         & 83.00       & 84.12      &  85.34    & 85.41     \\
            \textit{DomainNet}      & 33.28      & 35.75     &  37.74     & 37.83      \\
            \bottomrule
            \end{tabular}
    \end{minipage}
    \hspace{.3cm}
    % \hfill
    \begin{minipage}{.5\textwidth}
        \caption{ACC as a function of training epoch.}
        \label{tab:t3-1}
        \centering
             \begin{tabular}{ccccc}
             \toprule
             training epoch          & 120        & 180             & 240         & 300 \\ \midrule
             \textit{Digits}         & 80.10       & 83.46           & 85.34       & 85.38     \\
             \textit{DomainNet}      & 34.80       & 36.50            & 37.74        & 38.16         \\
             \textit{Office-Caltech} & 80.93      & 84.70            & 87.23       & 87.28     \\ \bottomrule
            \end{tabular}
    \end{minipage}
}
\end{table}

Table~\ref{tab:t2-1} shows the ACC of GRCL under various memory sizes per domain. The ACC benefits from a larger memory size.
Because larger domain memory provides more negative samples for contrastive learning, resulting in better self-supervised representation learning.
As the Office-Caltech dataset only has 100 training samples per domain, it is not applicable to do the ablation of memory sizes.

Table~\ref{tab:t3-1} shows the  ACC of GRCL with different training schedules. Because contrastive learning naturally benefits from longer training schedules~\cite{he2019momentum}, GRCL consistently got improvements of ACC as more training epochs are adopted.
But the performance gain soon saturates after 240 epochs.

% \section{Ablation Study}

\subsubsection{Importance of Gradient Regularization in GRCL} \label{sec:grcl-exp}

\begin{figure}[!h]
    \centering
    \includegraphics[width=1\linewidth]{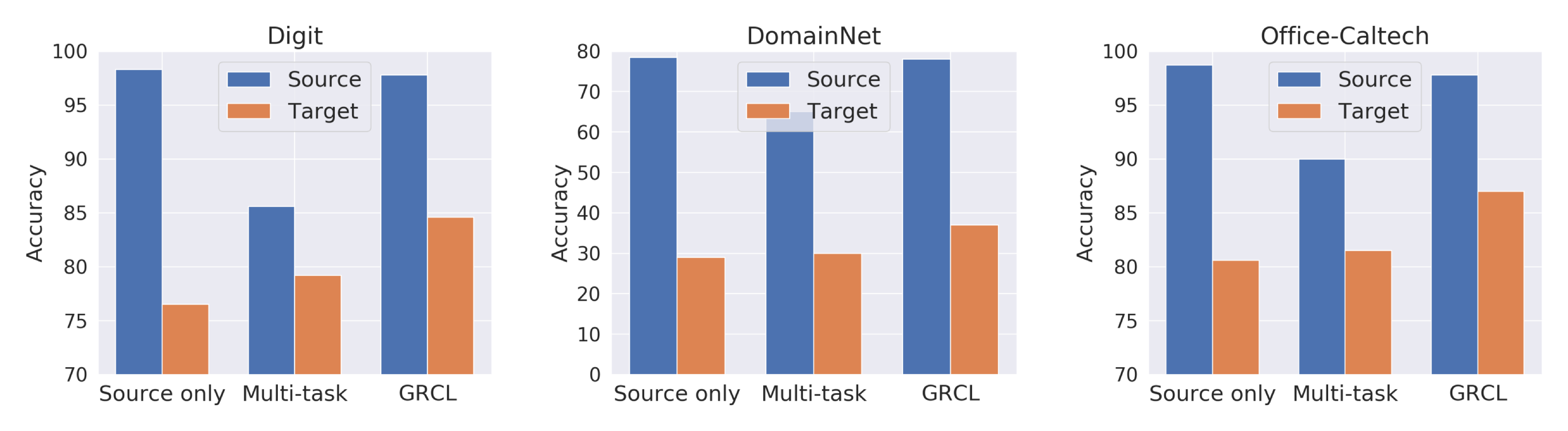}
    \caption{Comparison among Source only, Multi-task and GRCL. The performance on target represents the averaged accuracy over all different target domains.}
    \label{fig:source-target-acc}
\end{figure}
In this section, we evaluate the effect of gradient regularization of GRCL on one target domain setting.
We compare three different methods: (1) source only, which the model is supervised trained on source domain; (2) Multi-task, using multi-task objective of Eq.\ref{eq:contrast-da-multi-task}; (3) GRCL.
The $\lambda$ in Eq.\ref{eq:contrast-da-multi-task} uses the best value obtained via grid search on each target domain.
The SynNum, Clipart and DSLR are used as the source domain for Digits, DomainNet and Office-Caltech dataset respectively.
We report the averaged classification accuracy on the different target domains.
As shown in Fig.\ref{fig:source-target-acc},
% the source only method performs well on the source domain, but worse on the target domain, due to the domain gaps.
Multi-task improves the performance on the target domain over the baseline of source only method.
Because of the conflicts between cross-entropy loss and contrast loss,  Multi-task trained model sacrifices the performance of original source domain, which in turn hurts the discriminative power of image features of the target domain.
Benefiting from the gradient regularization, GRCL  enjoys the domain-invariant feature learning brought by contrast loss, and semantically discriminative feature provided by cross-entropy loss, which together helps it achieve the better classification accuracy on the target domain.
% The detailed results are shown in Table~\ref{tab:3}.

% \newpage

\section{Related Works}

\paragraph{Continual domain adaptation}
\cite{bobu2018adapting} adopts adversarial training to align different domain distributions and reuse self-labeled samples of previous domains to retain classification performance.
\cite{mancini2019adagraph} propose AdaGraph to encode inter-domain relationships with domain metadata (i.e., the viewpoint of an image captured), and utilize it produce domain-specific BN parameters.
However, existing methods have two limitations:
(1)\cite{bobu2018adapting} requires domain labels to do domain adversarial training and
\cite{mancini2019adagraph} requires additional metadata as prior to build the domain relation graph, which may not accessible in practice.
In contrast, GRCL leverages self-supervised learning to learn domain-invariant representation, which does not reacquire priors of domain-related information.
(2)\cite{bobu2018adapting} uses sample replay to avoid catastrophic forgetting.
However, the simple buffer-replay based methods still suffer from the negative transfer, as shown by~\cite{lopez2017gradient}.
In contrast, our method explicitly constraints domain adaptation learning on the new domain not to increase the loss on previous domains.
Thus it theoretically permits positive backward transfer that existing method~\cite{lopez2017gradient} does not support.

% \vspace{-5pt}

\paragraph{Contrastive learning}
has been a powerful self-supervised learning method to learn semantic image representation that can be transferred to downstream vision recognition tasks.
It utilizes a contrastive loss to encourage the model to embed similar images closer to each other while dissimilar images separate in the feature space.
Different contrastive learning methods vary in the strategy to generate positive and negative image pairs for each input.
For example, \cite{wu2018unsupervised} samples the images pairs from a memory bank, ~\cite{he2019momentum,chen2020improved} adopt a momentum encoder to generate the image pairs and \cite{tian2019contrastive,chen2020simple} produce the image pairs using the current batch of images.
In this work, we utilize the contrastive learning to jointly learn domain-invariant image representation and the gradient regularization is proposed to maintain the discriminative power of learned representation.

% \vspace{-5pt}

\paragraph{Continual learning}  addresses the catastrophic forgetting in a sequence of supervised learning tasks.
One representative technique is using episodic memory~\cite{lopez2017gradient,chaudhry2018efficient} to  store some training samples of old tasks, which are used to overcome catastrophic forgetting via constrained optimization.
In contrast to continual learning that considers a sequence of supervised learning tasks without domain shift problem, continual domain adaptation aims to solve a sequence unsupervised domain adaptation tasks under domain shifts.
Hence continual DA not only requires the learner to overcome catastrophic forgetting like lifelong learning dose but also needs the learner to adapt the novel target domain with varying data distributions.

\section{Conclusion}
This work studies the problem of continual DA, which is one major obstacle in the deployment of modern AI models.
We propose Gradient Regularized Contrastive Learning (GRCL) to joint learn both discriminative and domain-invariant representations.
At the core of our method, gradient regularization maintains the discriminative power of feature learned by contrastive loss and overcomes catastrophic forgetting in the continual adaptation process.
Our experiments demonstrate the competitive performance of GRCL against the state-of-the-art.

\section{Broader Impact}
Deep neural networks have achieved great success on many supervised learning tasks, where enormous data annotations are available and the data distribution does not change over time.
However, it remains challenging for deep neural networks to adapt to novel domains, whose data distributions change over time and the data annotations may not be available.
The positive impact of this work is providing a method that enables continually adapt to environmental changes by leveraging past learning experience.
At the same time, this work may increase the risk of data privacy because now more unlabeled data can be used to improve the AI models.
Recently, many privacy-preserving training methods have been explored to improve data privacy, such as federated learning and SecureML.
In the future, these privacy-preserving methods can be combined with a continual domain adaptation approach to improve AI models while preserving data privacy.

\bibliographystyle{unsrt}
\bibliography{refs}

\newpage

\begin{appendices}
% \section*{Appendix}

\section{Algorithm of Gradient Regularized Contrastive Learning} \label{a:algo-grcl}

% Algorithm \ref{algo:1} summarizes the training protocol of Gradient Regularized Contrastive Learning (GRCL).

\begin{algorithm}[!h]
	\caption{GRCL for Continual Domain Adaptation}
	 \label{algo:1}
	\begin{multicols}{2}
	\begin{algorithmic}
    \footnotesize
	    \Procedure{Train}{$\theta, \mathcal{D}_s, \{ \mathcal{D}_1, \dots \mathcal{D}_T \}$}
	    \State $\mathcal{M} \xleftarrow{} \{ \}$
        \State $\mathcal{B}_1 \xleftarrow{} \Call{BuildFeatureBank}{\theta, \mathcal{D}_1, \mathcal{M} }$
		\For {$t = 1, \dots, T$}
% 		\State $\mathcal{M}_{s \bigcup t} \xleftarrow{} \Call{UpdateMemory}{\theta, \mathcal{D}_s, \mathcal{D}_t }$
		\For { $(x, i)$ in  $\mathcal{D}_s \cup \mathcal{M} \cup  \mathcal{D}_t$}
		\State $g_t \xleftarrow{}  \nabla \mathcal{L}_{contrast}(\theta, \mathcal{B}_{t}) $
		\State $g_s \xleftarrow{}  \nabla \mathcal{L}_{ce}(\theta, \mathcal{D}_s)   $
		\State \Call{UpdateKey}{$x, i,  \mathcal{B}_{t}$}
		\State $(x', y') \sim\mathcal{M} $
		\State $g_{dm} \xleftarrow{}  \nabla \mathcal{L}_{ce}(\theta, (x', y')  ) $
		\State $\hat g_t \xleftarrow{}  \Call{Solve}{g_t, g_s, g_{dm}} $, see Eq.8
		\footnotesize
        \State $\theta \xleftarrow{} \theta - \eta \hat g_t $
		\EndFor
% 		\State \Call{Evaluate}{$\theta, \mathcal{D}_t^{test}$}, $\forall t \in [1, \dots T]$
		\State $\mathcal{B}_{t+1}  \xleftarrow{} \Call{BuildFeatureBank}{\theta, \mathcal{D}_t, \mathcal{M} }$
		\State $\mathcal{M}  \xleftarrow{} \Call{UpdateMemory}{\theta, \mathcal{D}_t, \mathcal{M} }$
		\EndFor
		\State \Return $\theta$
		\EndProcedure
	\end{algorithmic}
	\columnbreak
    \begin{algorithmic}
    \footnotesize
    \Procedure{UpdateMemory}{$\theta, \mathcal{D}_t, \mathcal{M}$}
    % \State $\mathcal{\hat D}_t \xleftarrow{}$ a subset of $\mathcal{D}_t$ with confident prediction
    \State $\mathcal{\hat D}_t \xleftarrow{}$ run $k$-means on $\mathcal{D}_t$
    \State $\mathcal{M} \xleftarrow{} \mathcal{M} \bigcup \mathcal{\hat D}_t$
    \State \Return $ \mathcal{M}$
    \EndProcedure

    \Procedure{BuildFeatureBank}{$\theta, \mathcal{D}_t, \mathcal{M}$}
    \State $\mathcal{B}_{t+1} = \{ \}$
    \For { $x$ in  $\mathcal{D}_s \cup \mathcal{M} \cup \mathcal{D}_t$}
    \State $ k = g(f(x; \theta))$
    \State $k = k / \| k \|_2^2$
    \State  $\mathcal{B}_{t+1}  \xleftarrow{} \mathcal{B}_t \bigcup \{ k\}$
    \EndFor
    \State \Return $ \mathcal{B}_{t+1} $
    \EndProcedure

    \Procedure{UpdateKey}{$x, i$}
        \State $k(x) \xleftarrow{} m \cdot k^{i-1}(x) + (1-m) k^i(x)$
    \EndProcedure
    \end{algorithmic}
    \end{multicols}
\end{algorithm}

\newpage

\section{Quadratic Program of Equation (8)} \label{a:proof}

\textit{Proof}
The optimization objective in the Equation (8) of the main paper is:
\begin{equation} \label{eq:a0}
    \begin{aligned}
    \min_{\hat g_t} \quad & \frac{1}{2} \| g_t -  \hat g_t \|_2^2 \\
    \text{subject to} \quad  & \langle \hat g_t,  g_s \rangle \geq 0 \\
     & \langle \hat g_t, g_{dm} \rangle \geq 0.
    \end{aligned}
\end{equation}
Replacing $ \hat g_t$ with $z$ and discarding the constant term of $g_t^{T}g_t $, Equation (\ref{eq:a0}) can be rephrased  as
\begin{equation} \label{eq:a1}
    \begin{aligned}
    \min_{z} \quad  & \frac{1}{2} z^{T}z - g_t^{T}z \\
    % \min_{z} \quad  & \frac{1}{2} z^{T}z - g_t^{T}z + \frac{1}{2} g_t^{T}g_t  \\
    \text{subject to} & \quad   Gz \geq 0,
    \end{aligned}
\end{equation}
where  $G = -(g_s, g_{dm}) \in \mathbf{R}^{ 2 \times P}$ ($P$ is the number of parameters in the model).
The Lagrangian of Equation (\ref{eq:a1}) can be written as :
\begin{equation} \label{eq:a2}
    \mathcal{L}(z, u) =  \frac{1}{2} z^{T}z - g_t^{T}z - u^{T}Gz,
\end{equation}
where $u \in \mathbf{R}^{2 \times 1} \geq 0$.
Defining the dual of Equation(\ref{eq:a2}) as:
\begin{equation} \label{eq:a3}
  Q(u) = \inf_{z} \mathcal{L}(z, u).
\end{equation}
We can  find the value $z^*$ that minimizes the $ \mathcal{L}(z, u) $ by setting the derivatives of $\nabla_{z} \mathcal{L}(z, u) $ to zero:
\begin{equation}
    \begin{aligned}
    %   &  \nabla_{z} \mathcal{L}(z, u)=0 \\
      & z^* = G^{T}u + g_t.
    \end{aligned}
\end{equation}
Equation (\ref{eq:a3}) can be simplified by substituting the value of $z^*$:
\begin{equation}
    Q(u) = -\frac{1}{2} u^{T}GG^{T}u - g_t^{T}G^{T}u.
\end{equation}
So the Lagrangian dual of Equation (\ref{eq:a0}) becomes
\begin{equation}
    \min_{u} \quad   \frac{1}{2} u^{T}GG^{T}u + g_t^{T}G^{T}u  \quad
    \text{subject to} \quad   u \geq 0.
\end{equation}

\end{appendices}

\end{document}